%% file: acl_latex.tex
\pdfoutput=1

\documentclass[11pt]{article}

\usepackage[final]{acl}

\usepackage{times}
\usepackage{latexsym}
\usepackage{amsmath,amssymb,amsfonts}
\usepackage[noend]{algpseudocode}
\usepackage{pifont}
\usepackage{wrapfig}
\usepackage{bm}	
\usepackage{bbm}
\usepackage{float}
\usepackage{newfloat}
\usepackage{array}
\usepackage{placeins}
\usepackage{listings}
\usepackage{nicefrac} 
\usepackage{graphicx}
\usepackage{bbding}
\usepackage{dsfont}
\usepackage{mathtools}
\usepackage{fontawesome}
\usepackage{makecell}
\usepackage{multirow}
\usepackage{ragged2e}
\usepackage{color, colortbl}
\usepackage{xcolor}
\usepackage{subcaption}
\usepackage{microtype}
\usepackage{hyperref}
\usepackage{url}
\usepackage{booktabs}
\usepackage[T1]{fontenc}

\usepackage[utf8]{inputenc}

\usepackage{microtype}

\usepackage{inconsolata}

\usepackage{graphicx}
\usepackage{makecell}

\newcommand{\method}{Router-Tuning}
\newcommand{\weight}{\boldsymbol{W}}
\newcommand{\expert}{\boldsymbol{E}}

\newcommand{\mask}{\boldsymbol{M}}
\newcommand{\x}{\boldsymbol{x}}
\newcommand{\y}{\boldsymbol{y}}

\newcommand{\F}{\boldsymbol{F}}
\newcommand{\eL}{\mathcal{L}}
\newcommand{\gr}{\rowcolor[gray]{.95}}

\newcommand{\gate}{\boldsymbol{G}}
\newcommand{\router}{\boldsymbol{R}}

\title{Router-Tuning: A Simple and Effective Approach for Dynamic Depth}

\author{Shwai He\textsuperscript{\rm 1}\space\space\space\space
Tao Ge\textsuperscript{\rm 2}\space\space\space\space
Guoheng Sun\textsuperscript{\rm 1}
\space\space\space
Bowei Tian\textsuperscript{\rm 1}
\space\space\space\space \\
\textbf{Xiaoyang Wang}\textsuperscript{\rm 2}\space\space\space\space
\textbf{Dong Yu}\textsuperscript{\rm 2}\\
    \textsuperscript{\rm 1}University of Maryland, College Park \space\space
    \textsuperscript{\rm 2}Tencent AI Lab, Bellevue, WA \\
    {\tt\small shwaihe@umd.edu}}

\begin{document}    
\maketitle
\begin{abstract}
The Mixture of Depths (MoD) was introduced to improve computational efficiency by dynamically skipping less important layers, reducing redundant computation while maintaining model capacity. Despite its promise, existing MoD approaches remain under-explored and face two main challenges: (1) \textit{high training costs due to the need to train the entire model along with the routers that determine which layers to skip}, and (2) \textit{performance degradation when important layers are bypassed}. 
In response to the first issue, we propose Router-Tuning, which fine-tunes only the routers on a small dataset, drastically reducing the computational overhead associated with full model training. 
For the second challenge, we investigate \method~across different architectures and granularities, demonstrating its effectiveness on Attention layers and MoE layers. This method preserves the model’s performance while significantly enhancing computational and memory efficiency. Extensive experiments demonstrate that our approach delivers competitive results while dramatically improving the computation efficiency, e.g., 21\% speedup and only a 0.2\% performance drop. The code is released at \url{https://github.com/CASE-Lab-UMD/Router-Tuning}.  
\end{abstract}

\input{sections/Introduction}

\input{sections/Relatedworks}

\input{sections/Method}
\input{sections/Experiments}

\section{Conclusion}
In this work, we investigate the dynamic depth mechanism from both design and training perspectives. We propose \method, which effectively implements dynamic depth by fine-tuning only a minimal number of parameters in just a few steps. 
Additionally, we explore Router-Tuning across a variety of modules and granularities to evaluate its effectiveness across a wide range of models and tasks. These advancements provide valuable insights and practical solutions for deploying dynamic depth and enhancing the efficiency of large language models. 
 
\input{sections/Limitations}

\bibliography{custom}

\end{document}

%% file: sections/Introduction.tex
\section{Introduction}
Large Language Models (LLMs) have shown promising performance across various domains \cite{openai2024gpt4technicalreport, geminiteam2024gemini, deepseekai2024deepseekv3technicalreport}. However, the continuous increase in model size leads to substantial computational costs in real-world applications, making computation reduction a critical research focus for improving LLM efficiency \cite{sun2024simpleeffectivepruningapproach, lin2024awqactivationawareweightquantization}.
A promising approach to this challenge is the Mixture of Depths (MoD) \cite{raposo2024mixtureofdepthsdynamicallyallocatingcompute}, which dynamically allocates computational resources for specific inputs. 
Instead of uniformly applying all layers to every input, MoD selectively activates only a subset of the model’s layers, skipping those deemed less important. This targeted activation significantly reduces computational overhead while maintaining performance. 

Despite its potential, current MoD methods are still underexplored and face several critical challenges. On the one hand,  the involvement of additional router networks, which decide which layers to skip, often requires extra extensive training: \citet{raposo2024mixtureofdepthsdynamicallyallocatingcompute} train the entire model from scratch while \citet{tan2024dlodynamiclayeroperation} performs costly continual training. This creates a significant barrier to efficiently integrating MoD with existing LLMs. Furthermore, most prior MoD implementations \cite{raposo2024mixtureofdepthsdynamicallyallocatingcompute, tan2024dlodynamiclayeroperation} have been applied to transformer blocks and MLP layers, which are sensitive to skipping. As a result, omitting important components often leads to significant performance degradation \cite{he2024matterstransformersattentionneeded}. 

These challenges prompt us to reflect on the two key questions: \textit{(1) How can we implement dynamic depth to improve efficiency without incurring excessive training costs? (2) How can we preserve model performance in the presence of dynamic depth?} 

To tackle the first challenge, we introduce \textit{Router-Tuning}, a novel method that fine-tunes only the router network without updating the backbone model’s parameters. As each router network is a lightweight, single-layer projector that accounts for less than 0.01\% of the total parameters, the training overhead is minimal and even significantly lower than that of parameter-efficient finetuning methods \cite{houlsby2019parameterefficienttransferlearningnlp, he2021towards} like LoRA \cite{hu2022lora}.  Router-tuning requires only a small-scale dataset and fewer training steps, eliminating the need for large-scale pretraining or extensive continual training. Meanwhile, by freezing the backbone, Router-Tuning contributes to mitigating catastrophic forgetting and better retaining the original model performance \cite{Houlsby2019ParameterEfficientTL, liu2024delvingparameterefficientfinetuningcode, qiao2024learn}. These properties make Router-Tuning a highly efficient and scalable solution for dynamic adaptation. 

To address the second challenge, we conduct a comprehensive investigation of target modules (e.g., Block, MLP, Attention, MoE), and various granularities (e.g., token and sequence).  
For dense transformer architectures, we propose \textit{Attention with Dynamic Depths}, which selectively applies dynamic depth to attention layers. By focusing on attention layers known to exhibit high redundancy \cite{he2024matterstransformersattentionneeded}, \method~not only preserves model accuracy but also alleviates computational and memory bottlenecks. 
In the case of Mixture-of-Experts (MoE) layers \cite{shazeer2017outrageouslylargeneuralnetworks, fedus2022switchtransformersscalingtrillion}, where efficiency is often hindered by the computational cost of activating multiple expert networks, we apply Router-Tuning at the expert level to enhance overall efficiency. 

Through extensive experiments, we demonstrate the effectiveness of our approach across multiple open-source language models, including Llama \cite{touvron2023llama2openfoundation}, Mistral \cite{jiang2023mistral7b}, Qwen \cite{bai2023qwentechnicalreport}, Deepseek-MoE \cite{dai2024deepseekmoe}, and OLMoE \cite{muennighoff2024olmoeopenmixtureofexpertslanguage}. Router-Tuning requires less than half an hour on an Nvidia RTX A6000, making it 1000 times faster than DLO \cite{tan2024dlodynamiclayeroperation}. \method~maintains a high percentage of the original model's performance while significantly reducing memory usage and accelerating inference, achieving, for example, a 21\% inference speedup with only a 0.2\% performance degradation. Furthermore, Router-Tuning can be seamlessly integrated with LoRA fine-tuning, further enhancing both efficiency and performance.  

In short, our contributions are as follows: 
\begin{itemize}
    \item We introduce \textit{Router-Tuning}, a lightweight method that fine-tunes only the router using a small dataset, effectively addressing the high computational cost of training the entire model with routers.
    
    \item We systematically investigate routing scopes, deployment granularities, and model architectures, demonstrating the effectiveness of Router-Tuning on Attention and MoE layers.
    
    \item Through comprehensive experiments, Router-Tuning achieves competitive performance while delivering substantial improvements in training and inference efficiency.
\end{itemize}

%% file: sections/Relatedworks.tex
\section{Related Work}
\paragraph{Layer Redundancy}
While increasing the depth of large language models has demonstrated promising performance across a wide range of tasks \cite{openai2024gpt4technicalreport, geminiteam2024gemini}, it also introduces layer redundancy \cite{gromov2024unreasonableineffectivenessdeeperlayers, he2024matterstransformersattentionneeded}, posing efficiency challenges. To address this issue, several approaches have been proposed to reduce model depth \cite{men2024shortgptlayerslargelanguage} while maintaining comparable performance. Surprisingly, removing redundant layers has been shown to preserve performance while significantly reducing memory and computational costs \cite{gromov2024unreasonableineffectivenessdeeperlayers, he2024demystifyingcompressionmixtureofexpertsunified,  he2024matterstransformersattentionneeded}. Specifically, 
\citet{gromov2024unreasonableineffectivenessdeeperlayers} suggest dropping continuous Transformer blocks, and 
\citet{he2024matterstransformersattentionneeded} propose fine-grained layer dropping to further improve the effectiveness of layer reduction.  
However, these static techniques fail to account for the varying complexity of different input sequences, where excessive layer removal can significantly degrade performance on more complex tasks. Instead of statically removing unimportant layers, our approach focuses on dynamically skipping these layers based on the specific inputs.

\paragraph{Dynamic Depth} 
Dynamic Depth, which allocates different layers based on the specific input, is an effective technique for accelerating inference while preserving performance \cite{han2021dynamic, han2022on}. Recent works primarily implement dynamic depth through two key methods: Early-Exit~\cite{bae2023fast, layerskip} and Mixture of Depths (MoD)~\cite{raposo2024mixtureofdepthsdynamicallyallocatingcompute}. Early-exit strategies terminate computation in later layers once sufficient confidence is achieved, effectively reducing redundant computations. In contrast, MoD offers greater flexibility by dynamically skipping less critical layers, enhancing adaptability and representational capacity.
Despite their advantages, both Early-Exit and MoD often involve significant training overhead. For instance, LayerSkip~\cite{layerskip} and MoD~\cite{raposo2024mixtureofdepthsdynamicallyallocatingcompute} require training models from scratch or extensive continual training, while Tan \textit{et al.}~\cite{tan2024dlodynamiclayeroperation} extends pre-trained model training over long schedules to achieve optimal performance.  To overcome these limitations, we propose \method, an efficient approach to dynamic layer skipping that requires minimal additional offline training, providing a more cost-effective solution.

%% file: sections/Method.tex
\section{Methodology}

\begin{figure*}[ht]
\centering
\makeatother\def\@captype{figure}\makeatother
	\centering
    \includegraphics[width=\textwidth]{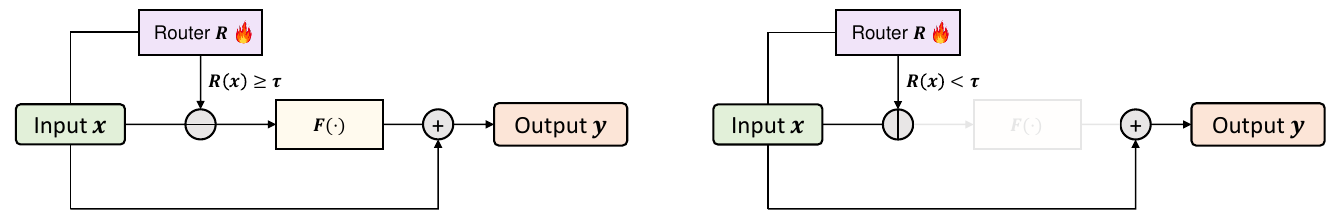}
        \vspace{-10pt} 
        \caption{
        \textbf{Overview of Router-Tuning. 
        } 
        Router-Tuning involves a trainable router to determine whether a given layer $\F(\cdot)$ (e.g., Attention and MLP) would be skipped. Inputs with routing scores lower $\router(\x)$ than the threshold $\tau$ are skipped, and only the router $\router$ is trainable in the whole model. }
\label{fig:overview}
\end{figure*}

In this section, we first review the challenges associated with deploying Mixture of Depths and then introduce \method, addressing the implementation of Mixture of Depths from both design and training perspectives. 

\subsection{Motivation}
\label{sec:motivation}
The Mixture of Depths (MoD) framework \cite{raposo2024mixtureofdepthsdynamicallyallocatingcompute}, which dynamically adjusts layer depth based on input complexity to enhance computational efficiency, was originally designed for integration during the pretraining phase, where transformer models are trained from scratch with MoD-enabled layers. More recently, \citet{tan2024dlodynamiclayeroperation} applied MoD to pretrained Llama models \cite{touvron2023llama2openfoundation} through continual training.
While these approaches have demonstrated promising results, \textbf{\textit{training with MoD remains computationally expensive and time-consuming, posing challenges for scalability and real-world deployment}}. A more efficient alternative is to apply MoD directly to existing pretrained models, followed by lightweight fine-tuning of a subset of parameters \cite{Houlsby2019ParameterEfficientTL, hu2022lora}, significantly reducing both computational costs and training time. 

On the other hand, MoD has typically been implemented at the transformer block level. However, \textbf{\textit{skipping entire transformer blocks has shown to be suboptimal to maintain the performance}}. 
Inspired by \citet{he2024matterstransformersattentionneeded}, we recognize that each transformer block contains layers of varying importance. Aggressively skipping entire blocks risks omitting critical layers, potentially degrading performance. Instead, skipping fine-grained layers offers a more effective strategy for preserving model accuracy.
Moreover, unlike blocks that generally share the same architecture, individual layers impose different computational costs. For instance, in dense transformer models, attention layers are particularly expensive, with computational complexity scaling quadratically with sequence length and additional memory needed for KV cache storage. In contrast, in MoE models, MLP layers hold the majority of the parameters, leading to substantial communication and computation overhead.

Building on these insights, we propose \method, a cost-effective formulation of MoD that achieves a favorable trade-off between performance and computational costs.

\subsection{Router-Tuning for Dynamic Depth}
In this part, we propose \method~to address the challenges outlined in Section \ref{sec:motivation}.
As illustrated in Figure \ref{fig:overview}, \method~incorporates an additional trainable router that determines whether to skip the layer. Specifically, \method~can be deployed in two levels: (1) \textbf{\textit{token-level}}, where layers are dynamically skipped for individual tokens, and (2) \textbf{\textit{sequence-level}}, where layers are dynamically skipped for the entire sequence. Given an input $\x \in \mathbb{R}^{L \times d}$, 
the router first computes an importance score for the input: 
\begin{equation}
\small
 \router(\x)_i = 
 \begin{cases}
 \text{GATE}(\x_i), & \text {Token-level} \\
 \text{GATE}(\frac{1}{L}\sum_{i=1}^L \x_i), & \text {Sequence-level} \\
 \end{cases}, 
\end{equation}
where $\router$ is a scoring router that assesses the importance score of the input, GATE is the gating function $\text{GATE}(\x) = \text{Sigmoid}(\weight \x)$. Based on the computed importance scores, we further apply a binarized mask $\mask$ to determine whether to skip a token or an entire sequence: 
\begin{equation}
     \mask = \begin{cases} 
    1, & \text { if } \router(\x) \ge \tau \\ 
    0, & \text { otherwise }\end{cases},
\end{equation}
where $\tau$ is the threshold. The score is set to zero for skipped inputs and one for retained inputs, ensuring stable outputs \cite{tan2024dlodynamiclayeroperation}.

To enable a differentiable and trainable binary decision process, we utilize the straight-through estimator (STE) \cite{bengio2013estimatingpropagatinggradientsstochastic}, which allows gradients to propagate through the binary selection mechanism via $\frac{\partial {\mask}}{\partial {\router}} = 1$. The final output of MoD is then computed as follows:
\begin{equation}
    \y = 
    {\mask} \odot \F(\x) + \x, 
 \label{eq_mask}
\end{equation}
where $\F$ denotes a given layer and $\y$ is the output. This formulation ensures that the router is fully trainable through the gradient calculations: 
\begin{equation}
    \frac{\partial \y}{\partial \weight} = \frac{\partial \y}{\partial {\mask}} \frac{\partial {\mask}}{\partial {\router}} 
    \frac{\partial \router}{\partial {\weight}}. 
\end{equation}
During inference, without the need for gradient calculations, we further enhance computational efficiency by completely bypassing computations for skipped inputs:
\begin{equation}
    \y = \begin{cases}
        {\F}(\x) + \x, & \text{ if } \router(\x) \geq \tau
        \\
        \x, & \text{otherwise} 
    \end{cases}. 
\end{equation}
This dynamic routing mechanism ensures that computation is performed only when necessary,
thereby enhancing the computational efficiency.

\subsection{Extension to Mixture of Experts}
Mixture of Experts (MoE) employs sparse activation, dynamically selecting expert networks for each input, which delivers promising performance in various tasks \cite{jiang2024mixtralexperts, dai2024deepseekmoe, muennighoff2024olmoeopenmixtureofexpertslanguage}. 
However, MoE also exhibits significant redundancy, allowing certain experts or layers to be skipped with minimal impact on performance \cite{lu-etal-2024-experts, he2024demystifyingcompressionmixtureofexpertsunified}. Building on this, we extend \method~to MoE layers by implementing dynamic skipping within each expert: 
\begin{equation}
    \hat \expert_i(\x) = \begin{cases}
        \expert_i(\x), & \text { if } \router(\x) \geq \tau \\
        0, & \text{otherwise} 
    \end{cases},  
\end{equation}
where $\expert_i$ denotes the $i$-th expert and $\hat \expert_i(\x)$ is the corresponding output denotes the corresponding output, bypassing the skipped tokens. 
Given a collection of $n$ experts, $\{\expert_1, \expert_2, \dots, \expert_n\}$, the overall output of the MoE layer is as follows: 
\begin{equation}
    \mathcal{K} = \mathrm{TopK}(\mathrm{Softmax}(\gate(\x)), k),
\end{equation}
\begin{equation}
    \y = \sum\nolimits_{i\in\mathcal{K}} {\gate(\x)_i \cdot \hat \expert_i(\x)},
\label{eq:moe}
\end{equation}
where $\mathcal{K}$ denotes the indices of the top-$k$ selected experts, and $\gate(\x)_i$ represents the selection score for the $i$-th expert. By dynamically skipping experts within each layer, \method~significantly reduces computation costs.

\subsection{Training Objectives}

Given the computationally intensive nature of training entire LLMs and the constraints of real-world computational resources, our goal is to implement dynamic depth while minimizing both computational costs and time overhead. To achieve this, we focus exclusively on fine-tuning the routers, as illustrated in Figure \ref{fig:overview}, thereby eliminating the need for costly full-model training.

Specifically, we optimize two training objectives: improving task-specific performance and lowering MoD capacity (the proportion of non-skipped inputs). On the one hand, \method~is designed to maintain the performance of the original model, which we enforce using the loss term $\eL_\text{task}$ during fine-tuning. On the other hand, the model is encouraged to skip more tokens or sequences (i.e., reduce MoD capacity) to enhance efficiency. To achieve this, we introduce another loss term $\eL_\text{MoD}$, which drives the model to reduce MoD capacity to a desired target sparsity level $s$, thereby lowering computational costs and accelerating inference. The overall training objective is as follows: 
\begin{equation}
\eL = \eL_\text{task} + \lambda \cdot \eL_\text{MoD},  
\end{equation}
\begin{equation}
\eL_\text{MoD} = \text{ReLU}(\|M\|_0 - s), 
\end{equation}
where $\eL$ represents the standard loss function (e.g., cross-entropy), while $\eL_\text{MoD}$ is an $l_0$-norm regularization term that reduces MoD capacity. The coefficient $\lambda$ acts as a scaling factor to balance the trade-off between task performance and efficiency. 

%% file: sections/Experiments.tex
\section{Experiment Setup}
\label{sec:setting}
\paragraph{Models}
We conduct experiments on Llama \cite{touvron2023llama2openfoundation, grattafiori2024llama3herdmodels}, Qwen \cite{bai2023qwentechnicalreport}, and Mistral \cite{jiang2023mistral7b} due to their competitive performance and widespread adoption. Additionally, we leverage OLMoE \cite{muennighoff2024olmoeopenmixtureofexpertslanguage} and Deepseek-MoE \cite{dai2024deepseekmoe} as the backbone to deploy \method~on the Mixture of Experts.  

\paragraph{Datasets} For the training dataset, we used Llama-Pro \cite{wu2024llamaproprogressivellama}, given it spanning general instruction, math, and code for the SFT process and offering a wealth of instruction data with varying complexity levels.
To evaluate model performance, we report normalized zero-shot or few-shot accuracy on the LM-Harness benchmark. The number of shots for each task is detailed in Table \ref{tab:shot}, which includes multiple tasks: ARC-C \cite{clark2018think}, BoolQ \cite{clark2019boolq}, HellaSwag \cite{zellers2019hellaswag}, MMLU \cite{hendrycks2021measuring}, OBQA \cite{mihaylov2018suit}, PIQA \cite{bisk2019piqa}, RTE \cite{wang2019glue}, WinoGrande \cite{ai2:winogrande} and GSM8K \cite{cobbe2021gsm8k}. The evaluation code is based on EleutherAI's LM Harness framework \cite{eval-harness}.

\begin{table}[ht]
    \centering
    \caption{\textbf{Experimental settings for evaluation tasks. } ``Norm'' refers to the normalization performed with respect to the length of the input.
    }
    \resizebox{\columnwidth}{!}{
    \setlength{\tabcolsep}{2pt}
    \begin{tabular}{lcc}
    \toprule
    \bf ~Task~ 
     & \bf ~Number of few-shot~ & \bf ~Metric~  \\
          \midrule
    BoolQ
    & 0 & Accuracy
    \\
    RTE
    & 0 & Accuracy
    \\
    OBQA
    & 0 & Accuracy (Norm)
    \\
    PIQA
    & 0 & Accuracy (Norm)
    \\
    MMLU
    & 5 & Accuracy
    \\
    WinoGrande
    & 5 & Accuracy
    \\
    GSM8K
    & 5 & Exact Match
    \\
    HellaSwag
    & 10 & Accuracy (Norm)
    \\
    ARC-C
    & 25 & Accuracy (Norm)
    \\
    \bottomrule
    \end{tabular}}
\label{tab:shot}
\end{table}

\paragraph{Hyperparameters} 
We set $\tau$ as 0.5, which corresponds to the midpoint of the sigmoid function. To ensure that training starts from dense models, 
we initialize $\weight$ to zero, ensuring that $\router(\x) \geq \tau$ initially, i.e., training from dense models. To achieve the desired MoD capacity, we perform a grid search over the learning rate from \{1e-5, 2e-5, 5e-5, 1e-4, 2e-4\} and the scale factor $\lambda$ from \{0, 0.1, 0.01, 0.001\}, respectively. 

\begin{table*}[htbp]
\renewcommand{\arraystretch}{0.9} 
\centering
\caption{
\textbf{Router-Tuning at different granularities.} We compare deployments on Attention, Block, and MLP layers. The number of skippable layers is capped at 16, with 50\% MoD capacity. SpeedUp denotes inference-time speedup.
}
\resizebox{\linewidth}{!}{
\begin{tabular}{l|l|c|cccccccc|c}
    \toprule
   \multicolumn{12}{c}{Llama-3-8B}
    \\ 
    \midrule
    Method~~
    & ~Granularity~ & ~Speedup~
    & ~ARC-C~ & ~BoolQ~ & ~HellaSwag~ & ~MMLU~ & ~OBQA~ & ~PIQA~ & ~RTE~ & ~WinoGrande~ & ~\underline{Avg.}~ \\
    \midrule
    Baseline 
    & \makecell{~~~~--}
    & $1.00\times$ 
    & 58.1 & 81.3 & 82.1 & 65.3 & 45.0 & 80.5 & 67.2 & 77.7 & \underline{69.7} \\ 
    \midrule
    \multirow{6}{*}{\method}
    & $\mathrm{Block}_{token}$
    & $\bf 1.24 \times$
    & 44.2 & 77.9 & 63.1 & 64.4 & 34.0 & 70.4 & 65.4 & 71.6 & \underline{61.4} 
    \\
    & $\mathrm{Block}_{seq}$
    & $\bf 1.26\times$  
    & 44.5 & 78.0 & 62.6 & 64.6 & 34.2 & 70.3 & 65.3 & 71.2 & \underline{61.3} \\ 
        \cmidrule(lr){2-12}
    & $\mathrm{MLP}_{token}$
    &  $1.05 \times$
    & 45.3 & 77.8 & 65.1 & 62.8 & 33.7 & 71.9 & 66.8 & 72.4 & \underline{62.0} \\ 
    & $\mathrm{MLP}_{seq}$
    & $1.06\times$ 
    & 45.1 & 77.7 & 65.4 & 62.4 & 33.4 & 71.6 & 66.4 & 72.1 & \underline{61.8} \\ 
    \cmidrule(lr){2-12}
    \gr
    \cellcolor{white}
    & $\mathrm{Attn}_{token}$
    & $\bf 1.18\times$  
    & 56.4 & 79.8 & 
    \bf 81.0 & \bf 65.3 & \bf 45.2 & 79.9 & 64.6 & 77.3 &  \underline{68.7} 
    \\
    \gr
    \cellcolor{white}   
    & $\mathrm{Attn}_{seq}$
    & $\bf 1.21\times$  
    & \bf 56.6 & \bf 80.5 & 
    80.7 & 65.1 & 44.6 & \bf 80.5 & \bf 69.7 & \bf 77.7 & \underline{\bf 69.4} \\ 
    \bottomrule
   \multicolumn{12}{c}{Llama-3-8B-Instruct}
    \\ 
    \midrule
    Method 
    & Granularity & Speedup 
    & ARC-C & BoolQ & HellaSwag & MMLU & OBQA & PIQA & RTE & WinoGrande & \underline{Avg.} \\
    \midrule
    Baseline 
    & \makecell{~~~~--}
    & $1.00\times$ 
    & 62.1 & 83.2 & 78.8 & 65.7 & 42.8 & 78.7 & 67.5 & 75.9 & \underline{69.3} 
    \\ 
    \midrule
    \multirow{6}{*}{\method}
    &  $\mathrm{Block}_{token}$
    &  $\bf 1.24 \times$
    & 44.6 & 80.9 & 54.1 & 60.2 & 31.2 & 64.8 & 67.7 & 65.1 & \underline{58.6} 
    \\
    & $\mathrm{Block}_{seq}$
    & $\bf 1.26\times$ 
    & 44.7 & 81.2 & 54.5 & 60.6 & 32.4 & 64.6 & 67.1 & 64.8 & \underline{58.7} 
    \\
    \cmidrule(lr){2-12}
    &  $\mathrm{MLP}_{token}$
    &  $1.05 \times$
    & 41.4 & 74.9 & 59.3 & 64.8 & 31.6 & 67.8 & 66.4 & 68.4 & \underline{59.3} 
    \\
    & $\mathrm{MLP}_{seq}$
    & $1.06\times$ 
    & 41.8 & 75.1 & 59.3 & 64.5 & 31.2 & 68.2 & 66.7 & 68.8 & \underline{59.5} 
    \\ 
    \cmidrule(lr){2-12}
    \gr
    \cellcolor{white}
    & $\mathrm{Attn}_{token}$
    & $\bf 1.18\times$  
    & 60.2 & 82.9 & 76.8 & \bf 65.8 & 42.6 & \bf 78.6 & 67.7 & 76.6 & \underline{68.9} 
    \\
    \gr
    \cellcolor{white}
    & $\mathrm{Attn}_{seq}$
    & $\bf 1.21\times$  
    & \bf 60.4 & \bf 83.3 & \bf 76.9 & 
    65.7 & \bf 43.0 & 78.2 & \bf 68.2 & \bf 76.9 & \bf \underline{69.1} \\ 
    \bottomrule
    \end{tabular}
}
\label{tab:main}
\end{table*}

\section{Main Results}

In this section, we evaluate the effectiveness of \method~on transformer architectures with the deployment details in Appendix \ref{sec:setting}.

\subsection{Performance of Router-Tuning}

\textbf{\method~achieves superior performance on Attention layers}
We first compare deploying \method~to different modules, e.g., Block, MLP, and Attention, as shown in Table \ref{tab:main}. 
Based on the observation that deeper layers are more redundant than shallow layers \cite{gromov2024unreasonableineffectivenessdeeperlayers, he2024matterstransformersattentionneeded}, we focus on deploying \method~to the deepest layers except the last one, leaving other layers unchanged. 

While previous studies have primarily explored layer dropping or skipping to Block and MLP layers \cite{bae2023fast, gromov2024unreasonableineffectivenessdeeperlayers}, skipping these modules significantly degrades performance when applied at either token or sequence level. In contrast, applying dynamic depth to Attention layers maintains the performance of original models, e.g., 69.4 v.s. 69.7 in Llama-3-8B. 
These findings reinforce our motivation to target Attention layers, and we utilize \method~on Attention layers by default unless stated otherwise. 

\begin{figure}[t]
\centering
\makeatother\def\@captype{figure}\makeatother
	\centering
\includegraphics[width=0.48\textwidth]{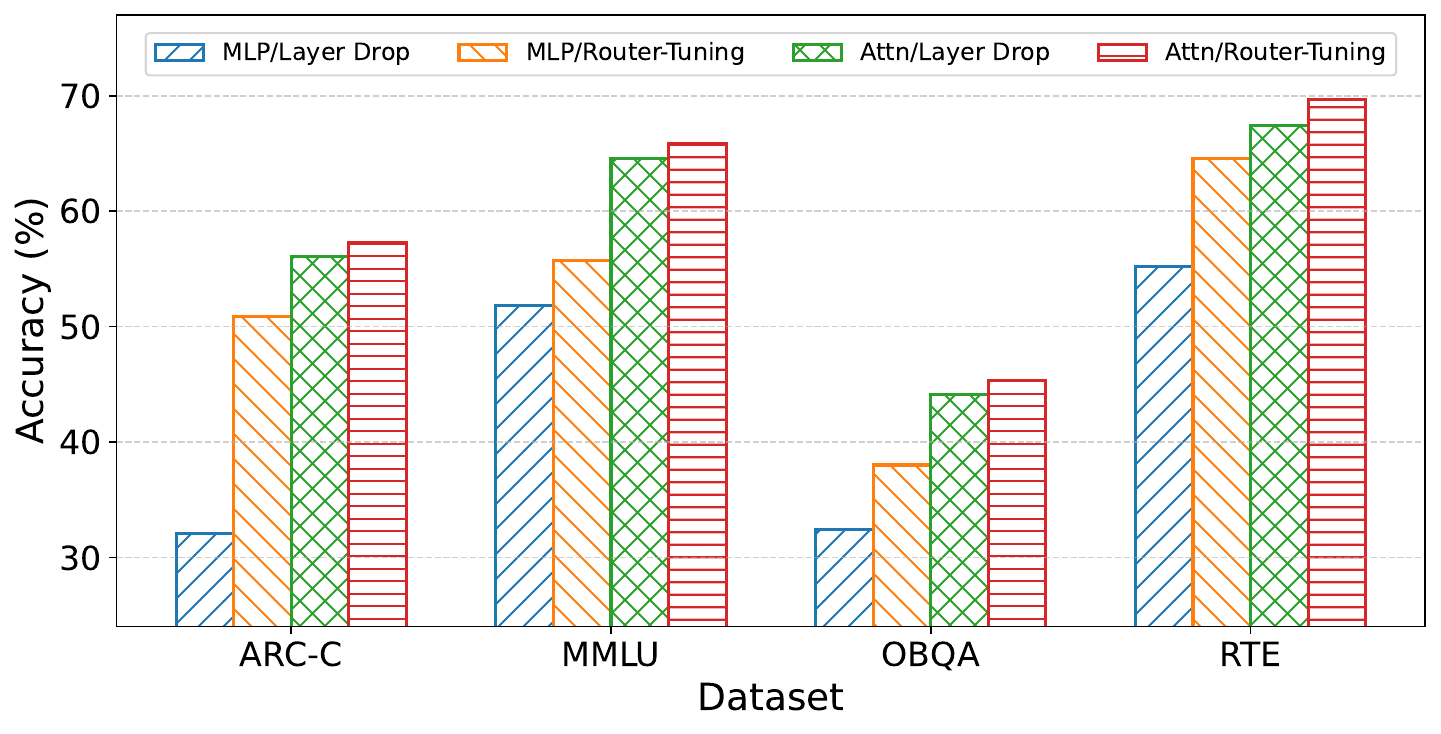}
    \vspace{-15pt}
\caption{\textbf{Comparison between \method~and Layer Drop} on MLP and Attention layers under a fixed 25\% overall skipping ratio. }
\label{fig:drop-mod-tasks}
\end{figure}

\begin{figure}[t]
\centering
\makeatother\def\@captype{figure}\makeatother
	\centering
\includegraphics[width=0.48\textwidth]{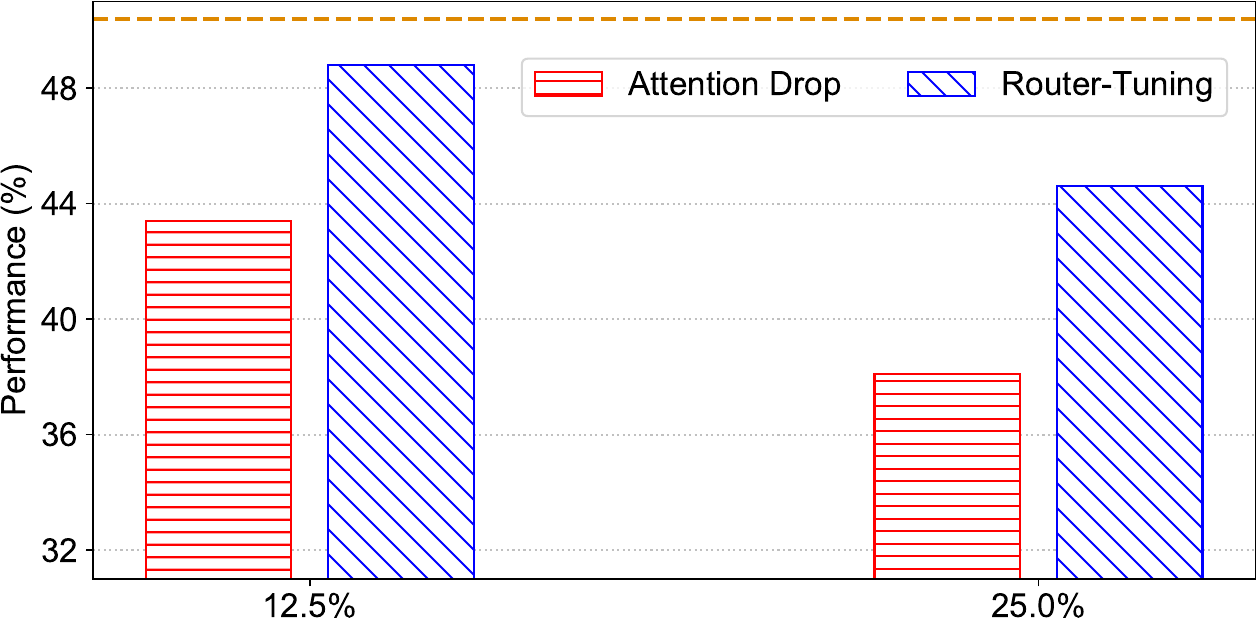}
    \vspace{-15pt}
\caption{\textbf{Comparison with Attention Drop} on GSM8K tasks under identical skipping ratios. }
    \vspace{-7pt}
\label{fig:drop-mod}
\end{figure}

\begin{figure}[t]
\centering
\makeatother\def\@captype{figure}\makeatother
	\centering
    \includegraphics[width=0.48\textwidth]{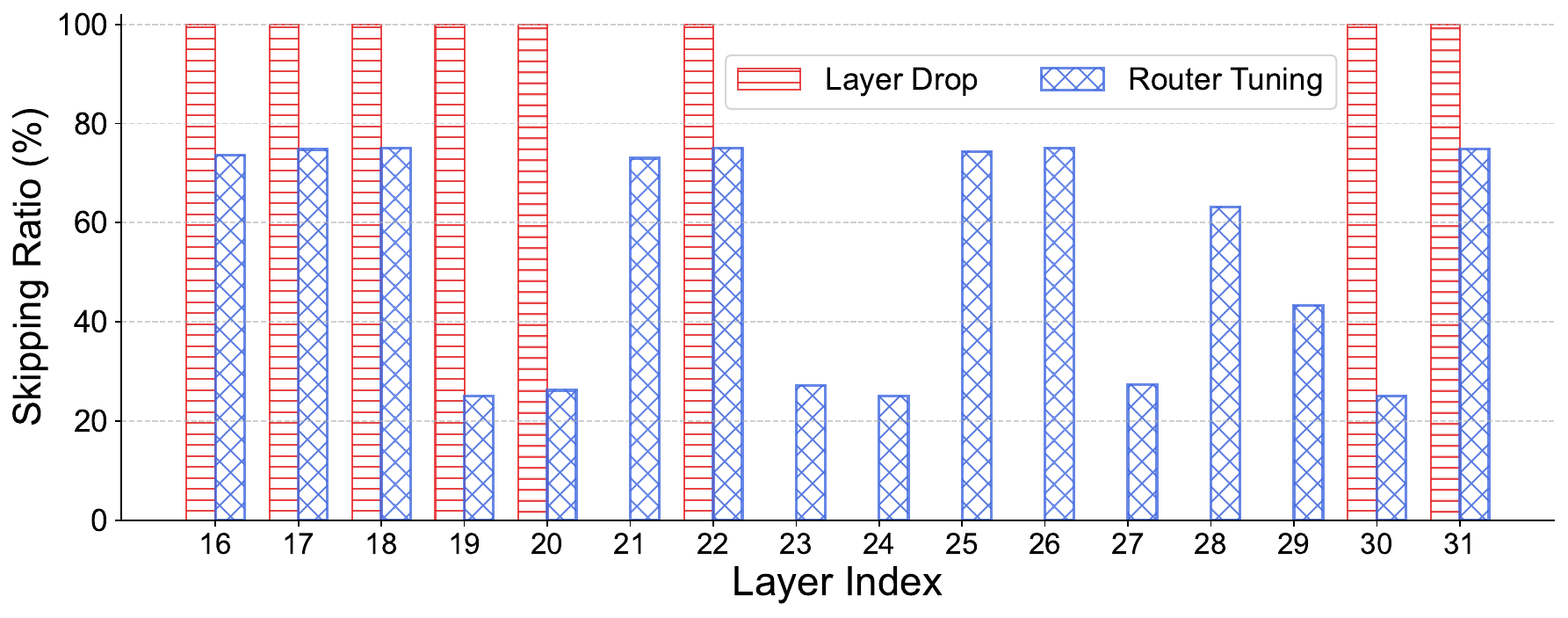}
    \vspace{-10pt}
        \caption{\textbf{Layer-wise Skipping Ratios} for Attention layers after Layer Drop and Router-Tuning. }
    \vspace{-15pt}
\label{fig:Skipping}
\end{figure}

\paragraph{\method~improves over static layer dropping}
While statically dropping attention layers \cite{he2024matterstransformersattentionneeded} has demonstrated promising performance, its static nature lacks flexibility and limits representational power. Here, we further investigate the improvements offered by the dynamic mechanism. Figure~\ref{fig:drop-mod-tasks} compares Router-Tuning with static Layer Drop \cite{he2024matterstransformersattentionneeded}, where Router-Tuning consistently achieves superior performance. For more complex tasks that are more sensitive to layer skipping, as shown in Figure \ref{fig:drop-mod}, we compare \method~with Layer Drop on attention layers (i.e., ``Attention Drop'') under the same computation budget, e.g., dropping 4 layers versus deploying MoD to 8 layers with 50\% capacity. Under the same skipping ratios, \method~significantly outperforms Attention Drop on the GSM8K benchmark \cite{cobbe2021gsm8k}, e.g., 6.5\% when the skipping ratio is 25.0\%. In Figure \ref{fig:Skipping}, we further visualize the layer-wise skipping ratios of MoD versus Attention Drop. Unlike static approaches that permanently remove certain layers, \method~maintains the utilization of all layers by adaptively distributing skipping ratios across them. This flexible allocation strategy contributes to improved performance. 

\paragraph{\method~outperforms dynamic skipping methods}
Table~\ref{tab:dynamic-skipping} compares \method~with dynamic skipping baselines, including DLO~\cite{tan2024dlodynamiclayeroperation} and Skip Transformer~\cite{peroni2024skip}. Following the setup of baselines, we conduct the comparison on the token level for MLP layers. 
\method~consistently outperforms these methods across most tasks, despite operating under the router-only training constraint. On the one hand, \method~freezes the model backbones, which contributes to avoiding the risks of catastrophic forgetting \cite{Houlsby2019ParameterEfficientTL, liu2024delvingparameterefficientfinetuningcode, qiao2024learn}. Additionally, \method~is trained end-to-end without being constrained by precomputed labels (e.g., token-level similarity scores in DLO) or stochastic gating mechanisms. This design enables more flexible and stable learning, while preserving the pretrained capabilities of the backbone.

\begin{table}[h]
\centering

\vspace{-5pt}
\caption{\textbf{Performance comparison against dynamic dropping baselines}, including DLO \cite{tan2024dlodynamiclayeroperation} and Skip Transformer \cite{peroni2024skip}. }
\vspace{-8pt}
\label{tab:dynamic-skipping}
\resizebox{\linewidth}{!}{
\begin{tabular}{l|cccc|c}
\toprule
\textbf{Method} & ARC-C & HellaSwag & MMLU & WinoG & \textbf{Avg.} \\
\midrule
DLO              & 44.5 & 64.2 & 62.1 & 71.3 & \underline{60.5} \\
Skip Transformer & 44.7 & 64.4 & 62.4 & 71.5 & \underline{60.8} \\
\method    & \textbf{45.3} & \textbf{65.1} & \textbf{62.8} & \textbf{72.4} & \underline{\textbf{61.4}} \\
\bottomrule
\end{tabular}
}
\end{table}

\subsection{Efficiency Improvements}

In this part, we measure the efficiency in both training and inference, focusing on computational and memory usage.

\begin{table}[h]
\centering
\caption{\textbf{Comparison of training strategies} of achieving dynamic layer skipping. The training time for MoD is left blank as it was not conducted on LLaMA-3-8B. }
\vspace{-5pt}
\label{tab:training-strategy}
\resizebox{\linewidth}{!}{
\begin{tabular}{l|c|c|c|c|c}
\toprule
\textbf{Method} & \textbf{Target Modules} & \textbf{Granularity} & \textbf{Training Stage} & \textbf{Trainable} & \textbf{Training Time} \\
\midrule
MoD~\cite{raposo2024mixtureofdepthsdynamicallyallocatingcompute} & Block & Token & Pretraining & Full Model & -- \\
DLO~\cite{tan2024dlodynamiclayeroperation} & MLP & Token & Continual Pretraining & Full Model & 36h on NVIDIA A100 \\
\textbf{Router-Tuning} & Block / MLP / Attn & Token / Sequence & Finetuning & Router & 15m on NVIDIA A6000 \\
\bottomrule
\end{tabular}
}
\label{tab:strategy}
\vspace{-5pt}
\end{table}

\paragraph{Training Efficiency}
The training efficiency of our method lies in two perspectives: trainable parameters and training steps. Since the router projects the input from dimension $d$ to 1, the number of trainable parameters is $d \times 1$ per layer, and the total number of trainable parameters is fewer than 0.01\% of the whole model. 
Additionally, Router-Tuning only requires a few steps, which is verified in Section \ref{sec:dataset}. Consequently, as shown in Table~\ref{tab:training-strategy}, Router-Tuning can be completed in under 15 minutes on a single NVIDIA A6000 GPU—over 1000 times faster than DLO~\cite{tan2024dlodynamiclayeroperation}, which performs large-scale training on full models and takes 36 hours on NVIDIA RTX A100 GPUs.

\paragraph{Inference Speedup} 
We also evaluate the run-time speed improvements achieved with \method. The inference speed is measured throughout the entire generation process, including prefilling and generation. To ensure that our results accurately reflect the performance gains, we adhere to two key principles: (1) all operations are performed on a single Nvidia RTX A6000 Ada GPU, eliminating any communication overhead from multi-GPU setups; and (2) we set the maximal sequence length as 2048 and increase batch sizes to fully utilize the GPU for each model. 

As shown in Table~\ref{tab:main}, skipping attention layers yields a more substantial speedup than skipping MLP layers, which is primarily due to the quadratic complexity of the attention mechanism and the memory overhead associated with KV-cache \cite{zhang2023ho, Singhania2024LokiLK, he2024matterstransformersattentionneeded}. 
On the other hand, different granularities contribute to different levels of speedup. This variation is primarily due to attention layers, where fine-grained token-level MoD introduces differing token lengths within a batch, necessitating padding operations to standardize sequence lengths. 
Instead, the speedups at the sequence level surpass those at the token level, with \method~achieving a 21\% improvement in inference speed. Therefore, we set \method~on the sequence level for attention layers as the default setting.

\paragraph{KV Cache} 
The KV cache stores intermediate representations of attention layers, accelerating inference by eliminating redundant computations but incurring substantial memory overhead. Our approach, which selectively skips attention layers, significantly reduces KV cache size—for instance, achieving an 8GB reduction when processing an input sequence of length 2048 with a batch size of 64 on Llama-3-8B. In contrast, DLO \cite{tan2024dlodynamiclayeroperation} operates exclusively on MLP layers and retains the full KV cache, providing no memory savings.

\begin{table*}[t]
    \centering
\caption{\textbf{Ablation study of \method~across multiple model architectures and scales}, highlighting its robustness and consistent improvements. }
\vspace{-5pt}
    \label{tab:models}
    \resizebox{\linewidth}{!}{
    \begin{tabular}{l|c|c c c c c c c c | c}
        \toprule
        Models~~
        & ~Speedup~
        & ~{OBQA}~ & 
        ~{PIQA}~ & ~{RTE}~ & 
        ~{WinoGrande}~ & 
        ~{BoolQ}~ & 
        ~{ARC-C}~ & 
        ~{HellaSwag}~ & 
        ~{MMLU}~ & ~~\underline{Avg.}~~ \\
        \midrule
        Llama-2-13B~~      
        & $1.00\times$    
        & 45.2 & 80.5 & 65.0 & 76.2 & 80.7 & 59.4 & 82.2 & 54.6 & \underline{68.0} \\
        \gr
        w/Router-Tuning~~         
        & $\bf 1.22\times$    
        & \bf 45.4 
        & \bf 80.6 
        & \bf 64.6 
        & \bf 76.2 
        & \bf 80.5 
        & \bf 59.3 
        & \bf 82.2 
        & \bf 54.7 
        & \bf \underline{67.9} \\
        \midrule
        Qwen-2.5-14B~~     
        & $1.00\times$    
        & 45.6 & 82.2 & 79.1 & 80.4 & 85.3 & 67.2 & 84.3 & 79.7 & \underline{75.5} \\
        \gr
        w/Router-Tuning~~         
        & $\bf 1.18\times$    
        & \bf 45.4 
        & \bf 82.4 
        & \bf 77.9 
        & \bf 78.5 
        & \bf 85.0 
        & \bf 66.2 
        & \bf 83.8 
        & \bf 78.0 
        & \bf \underline{74.7} \\
        \midrule
        Qwen-2.5-7B~~     
        & $1.00\times$    
        & 47.2 & 79.6 & 81.2 & 76.6 & 84.6 & 63.7 & 80.2 & 74.1 & \underline{73.4} \\
        \gr
        w/Router-Tuning~~       
        & $\bf 1.19\times$    
        & \bf 47.0 
        & \bf 80.1 
        & \bf 76.9 
        & \bf 76.1 
        & \bf 83.2 
        & \bf 62.3 
        & \bf 79.8 
        & \bf 73.3 
        & \bf \underline{72.3} \\
        \midrule
        Mistral-7B~    
        & $1.00\times$    
        & 44.4 & 82.2 & 68.2 & 79.0 & 82.2 & 60.6 & 83.2 & 62.4 & \underline{70.3} \\
        \gr
        w/Router-Tuning~
        & $\bf 1.24\times$    
        & \bf 44.0 
        & \bf 81.8 
        & \bf 67.6 
        & \bf 78.2 
        & \bf 81.7 
        & \bf 59.9 
        & \bf 82.6 
        & \bf 61.8 
        & \bf \underline{69.7} \\
        \bottomrule
    \end{tabular}}
\end{table*}

\section{Ablation Studies}

\paragraph{Compatibility across different models} 
Since \method~can be seamlessly integrated into pretrained language models, we extend our evaluation to diverse architectures, including Llama-2 \cite{touvron2023llama2openfoundation}, Mistral \cite{jiang2023mistral7b}, and Qwen2.5 \cite{bai2023qwentechnicalreport}, covering a wide range of model sizes. As shown in Table \ref{tab:models}, we deploy \method~in half of the attention layers while maintaining the total MoD capacity at 50\%. Across all models, \method~preserves performance compared to their original counterparts, demonstrating its effectiveness and adaptability across different architectures.
\begin{figure}[h]
\centering
\makeatother\def\@captype{figure}\makeatother
	\centering
\includegraphics[width=\linewidth]{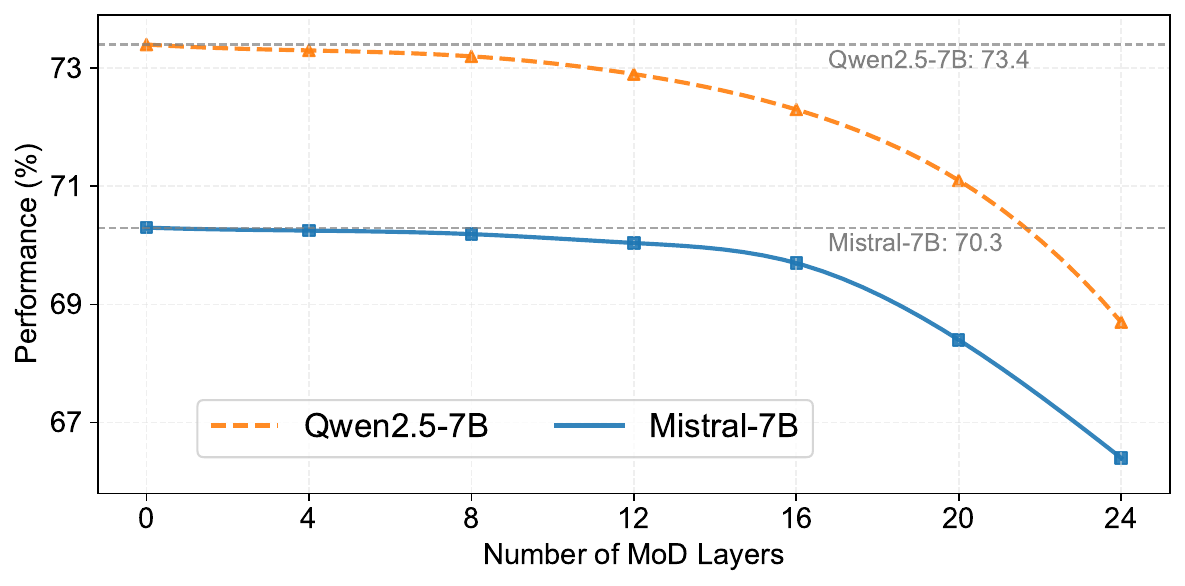}
            \vspace{-10pt}
\caption{Ablation study on the impact of varying the number of MoD layers on overall model performance.}
\label{fig:models}
\end{figure}

\begin{figure}[h]
\centering
\makeatother\def\@captype{figure}\makeatother
	\centering
\includegraphics[width=0.48\textwidth]{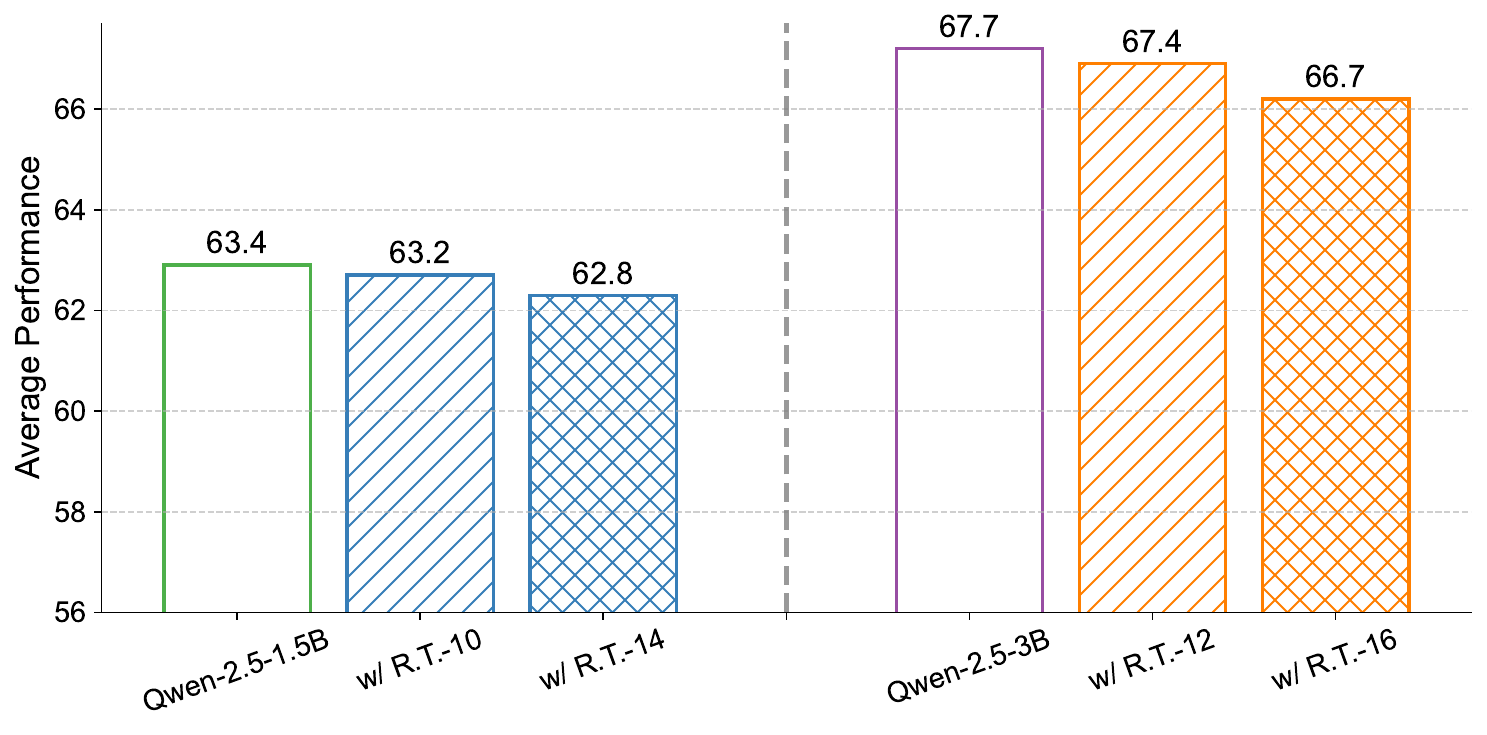}
    \vspace{-10pt}
\caption{\textbf{Performance of Router-Tuning on small models}, where ``R.T.'' denotes Router-Tuning and the postfix ``-$n$'' indicates that MoD is applied to $n$ layers.}
    \vspace{-5pt}
\label{fig:small_models}
\end{figure}

\paragraph{Impact of number of MoD layers}
\label{par:dataset}
In the main experiments, we deployed half of the layers with MoD. In Figure \ref{fig:models}, we further explore the effect of the number of MoD layers. Our results indicate that applying MoD to up to half of the attention layers still maintains comparable performance. A similar trend is observed in Figure~\ref{fig:small_models} for smaller models. 
However, when further increasing the number of MoD~layers, performance starts to degrade. We attribute this decline to the transformation of important shallow layers, which negatively impacts overall performance \cite{men2024shortgptlayerslargelanguage, he2024demystifyingcompressionmixtureofexpertsunified, he2024matterstransformersattentionneeded}. Therefore, preserving the density of shallow layers while applying MoD to deeper layers ensures its effectiveness. 

\paragraph{Influence of Training Dataset}
\label{sec:dataset}
In Table \ref{tab:dataset}, we next examine the impact of using different training datasets for \method. We consider a variety of datasets, including Alpaca \cite{alpaca}, Evol-Instruct \cite{xu2023wizardlmempoweringlargelanguage}, ShareGPT \cite{zheng2023judging}, and Llama-Pro \cite{wu2024llamaproprogressivellama}. Since \method~only fine-tunes the routers while keeping the backbone of the language models intact, changes in the training dataset do not significantly impact performance. However, Llama-Pro, which incorporates diverse training data from various domains, provides slightly better performance due to its broader data coverage. 

\begin{figure}[t]
\centering
\makeatother\def\@captype{figure}\makeatother
	\centering
\includegraphics[width=0.48\textwidth]{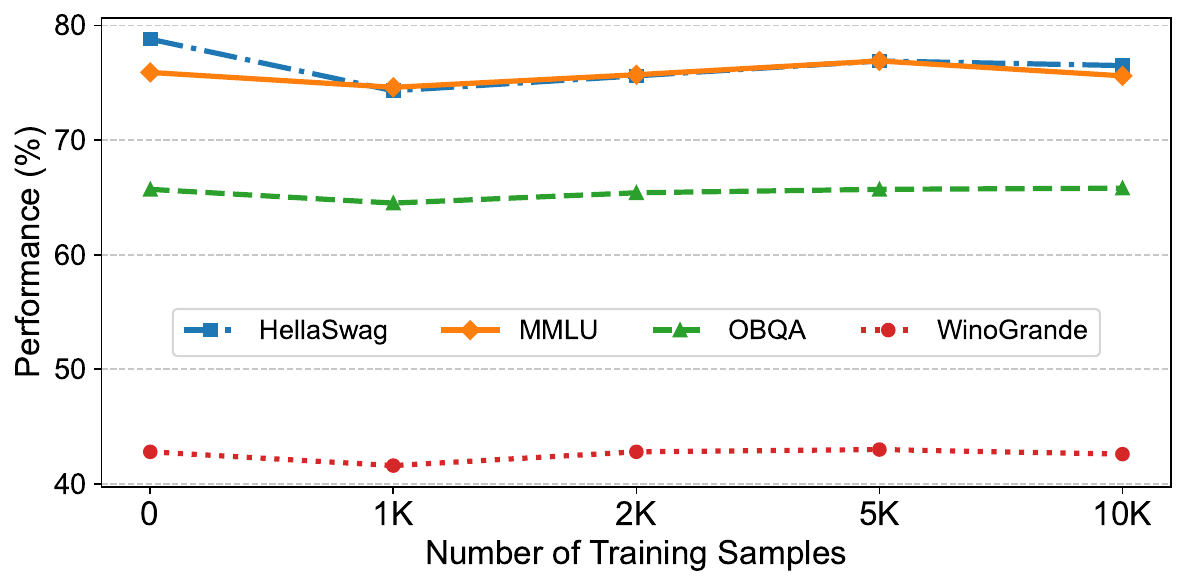}
    \vspace{-16pt}
\caption{Effect of varying the number of training samples on performance. } 
\label{fig:samples}
\end{figure}

On the other hand, due to the small number of trainable parameters, \method~does not require a large amount of training samples. As shown in Figure \ref{fig:samples}, MoD layers are initially dense-activated and then sparsified. Although the initial sparsification steps lead to a drop in performance, subsequent Router-Tuning facilitates performance recovery. Notably, just 5K training samples are sufficient to effectively train the routers. 

\begin{table}[ht]
    \centering
\caption{\textbf{Effectiveness across different training datasets}, where Router Tuning demonstrates robustness to the varying datasets. }
    \vspace{-6pt}
    \resizebox{\columnwidth}{!}{
    \setlength{\tabcolsep}{2pt}
    \begin{tabular}{l|cccc|c}
    \toprule
     Dataset~ & HellaSwag & ~MMLU~ & ~OBQA~ & ~WinoGrande~ 
     & \underline{
     Avg. 
     }  
    \\
    \midrule
    Baseline~ 
    & 82.1  & 65.3 & 45.0 & 77.7 & 
    \underline{67.5} 
    \\
    \midrule
    Alpaca
    & 79.8  & 62.2 & 43.8 & 77.4 & 
    \underline{65.8}  
    \\
    Evol-Instruct    
    & 80.4  & 64.0 & 44.4 & 77.6 & 
    \underline{66.6}  
    \\
    ShareGPT
    & 80.6  & 63.3 & \bf 45.4 & 76.7 & 
    \underline{66.5}  
    \\
    Llama-Pro
    & \bf 80.7  & \bf 65.1 & 44.6 & \bf 77.7 & 
    \underline{\bf 67.0}  
    \\
    \bottomrule
    \end{tabular}}
        \label{tab:dataset}
\end{table}

\section{MoE and LoRA Integration}
\begin{table*}[t]
    \centering
    \caption{\textbf{Performance of \method~on Mixture of Experts}, where we take Expert Drop \cite{he2024demystifyingcompressionmixtureofexpertsunified}  as the baseline of static dropping for comparison. }
    \vspace{-5pt}
    \label{tab:moe}
    \resizebox{\linewidth}{!}{
    \begin{tabular}{l|c|c c c c c c c c | c}
        \toprule
        Models~~
        & ~SpeedUp~ & ~{OBQA}~ & 
        ~{PIQA}~ & ~{RTE}~ & 
        ~{WinoGrande}~ & 
        ~{BoolQ}~ & 
        ~{ARC-C}~ & 
        ~{HellaSwag}~ & 
        ~{MMLU}~ & ~~\underline{Avg.}~~ \\
        \midrule
        DeepSeek-MoE      
        & $1.00\times$  
        & 43.6 & 80.5 & 62.8 & 73.4 & 72.4 & 52.7 & 79.9 & 44.5 & \underline{63.7} \\
        w/Expert Drop 
        & $\bf 1.11\times$  
        & 42.2
        & 80.2 & 59.9 & 70.0 & 74.0 & 48.1 & 75.6 & 38.9 & \underline{61.1} \\
        \gr
        w/Router-Tuning
        & $\bf 1.10\times$
        & \bf 43.2 
        & \bf 80.4 
        & \bf 61.2 
        & \bf 71.4 
        & \bf 72.1 
        & \bf 50.8 
        & \bf 77.3 
        & \bf 42.8 
        & \bf \underline{62.4} \\
        \midrule
        OLMoE   
        & $1.00\times$  
        & 45.6 & 80.1 & 53.7 & 71.2 & 74.7 & 54.5 & 79.4 & 52.5 & \underline{64.0} \\
        w/Expert Drop 
        & $\bf 1.13\times$
        & 34.0 & 66.6 & 51.6 & 59.3 & 67.6 & 39.0 & 40.5 & 42.7 & \underline{50.2} \\
        \gr
        w/Router-Tuning
        & $\bf 1.12\times$
        & \bf 40.4 
        & \bf 76.2 
        & \bf 53.2 
        & \bf 70.2 
        & \bf 71.3 
        & \bf 52.0 
        & \bf 77.9 
        & \bf 50.1 
        & \bf \underline{61.4} \\
        \bottomrule
    \end{tabular}}
\end{table*}

In this section, we further explore the integration of \method~with other architectures and training techniques. First, we implement \method~on mainstream MoE architectures. Then, we combine \method~with LoRA fine-tuning to enhance both efficiency and performance.
\begin{table*}[t]
    \centering
    \renewcommand{\arraystretch}{0.95} 
    \caption{\textbf{Effectiveness of \method~integrated with LoRA finetuning}, compared to deploying \method~and LoRA separately. }
    \vspace{-5pt}
\resizebox{\linewidth}{!}{ \setlength{\tabcolsep}{2pt}
    \begin{tabular}{l|c|c|cccccccc|c}
    \toprule
        Model~~ &
        ~~Method~~ & 
        ~~SpeedUp~~ & 
        ~~~{OBQA}~~ & 
        ~~{PIQA}~~ & 
        ~~{RTE}~~ & 
        ~~{WinoGrande}~~ & 
        ~~{BoolQ}~~ & 
        ~~{ARC-C}~~ & 
        ~~{HellaSwag}~~ & 
        ~~{MMLU}~~ & ~~~\underline{Avg.}~~~ 
        \\
    \midrule
        \multirow{4}{*}{Llama-3-8B~~} 
        & Baseline         
        & $1.00\times $
        & 45.0               & 80.5          & 67.2         & 77.7  & 81.3  & 58.1  & 82.1  & 65.3  
        & \underline{69.6}  \\ 
        \cmidrule(lr){2-12} 
        & R.T.                        
        & $1.21\times$
        & 44.6               & 80.5          & \bf 69.7         & 77.7  & 80.7  & 56.6  & 80.7  & 65.1  
        & \underline{69.5}  \\ 
        & LoRA                             
        & $1.00\times$
        & 46.6               & 82.0          & 68.0         & \bf 77.9  & \bf 83.9  & \bf 61.8  & 81.6  & \bf 65.9  
        & \bf \underline{71.0}  \\
        \gr
        \cellcolor{white}
        & LoRA + R.T.
        & $1.21\times$
        & \bf 47.2          
        & \bf 82.2          
        & 67.4         
        & 77.8  & \bf 83.9  & 61.5  & \bf 81.7  & 65.8  
        & 
        \underline{70.9}  \\ 
    \midrule
        \multirow{4}{*}{Mistral-7B} 
        & Baseline       
        & $1.00\times$
        & 44.4               & 82.2          & 68.2         & 79.1  & 82.2  & 60.6  & 83.2  & 62.4  
        & \underline{70.3}  \\ 
        \cmidrule(lr){2-12}
        & R.T.                          
        & $1.24\times$
        & 44.2               & 81.9          & 68.5         & 78.6  & 81.7  & 60.4  & 82.5  & 61.8  
        & \underline{70.0}  \\ 
        & LoRA              
        & $1.00\times$
        & 45.2               & 83.0          & \bf 68.9         & \bf 79.4  & \bf 84.7  & 60.7  & \bf 83.7  & 62.8  & 71.1  \\ 
        \gr
        \cellcolor{white}
        & ~~LoRA + R.T.~~
        & $1.24\times$
        & \bf 45.7               & \bf 83.1          & 68.7         & 79.3  & 84.3  & \bf 60.9  & 83.4  & \bf 62.9  
        & \bf \underline{71.2}  \\ 
    \bottomrule
        \end{tabular}}
        \vspace{-6pt}
    \label{tab:mod_lora}
\end{table*}

\paragraph{\method~on MoE}
The Expert's redundancy in MoE has been widely demonstrated in recent works \cite{lu-etal-2024-experts, he2024demystifyingcompressionmixtureofexpertsunified}, e.g., the model still maintains comparable performance after removing certain layers. 
Therefore, we further extend \method~to MoE, where we take OLMoE \cite{muennighoff2024olmoeopenmixtureofexpertslanguage} and DeepSeek-MoE \cite{dai2024deepseekmoe} as the backbones and equip each Expert network with \method. Since these models deploy MoE at the token level, we directly apply the token-level \method~to these models. Here, we compare \method~with Expert Drop~\cite{he2024demystifyingcompressionmixtureofexpertsunified} that statically drops less important experts. To ensure a fair comparison, we fix the overall skipping ratio of \method~to 25\%, and drop the bottom 25\% of experts globally by importance score in the Expert Drop baseline. 

\begin{figure}[h]
\centering
\makeatother\def\@captype{figure}\makeatother
	\centering
\includegraphics[width=0.48\textwidth]{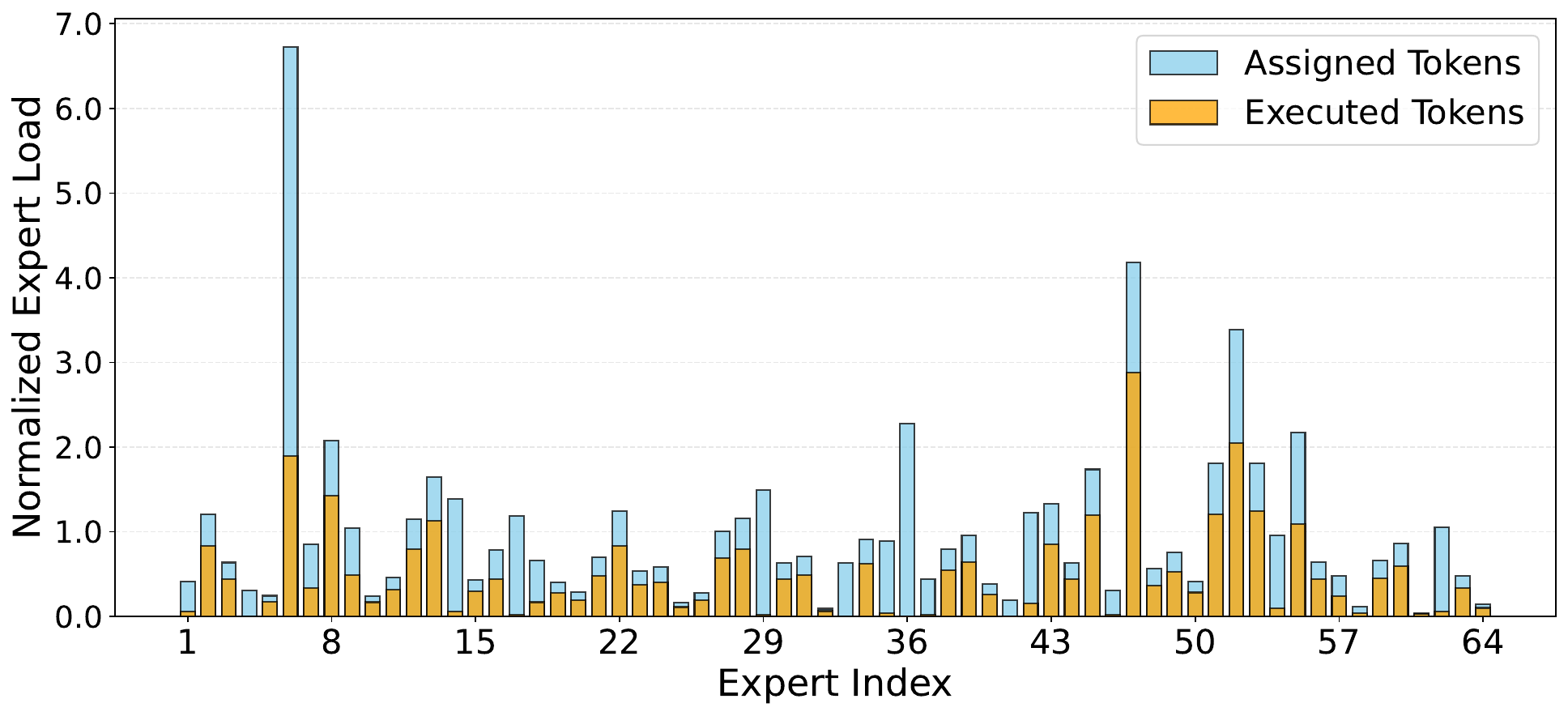}
    \vspace{-15pt}
\caption{
\textbf{Normalized expert load} before and after applying router-based filtering, denoted as ``Assigned Tokens'' and ``Executed Tokens'', respectively. Expert load values are normalized by the mean number of assigned tokens. 
}
\label{fig:experts}
\end{figure}

In Table \ref{tab:moe}, instead of removing a subset of experts like Expert Drop, 
\method~maintains the potential of all experts, which contributes to a superior performance. 
In Figure~\ref{fig:experts}, we further investigate how Router-Tuning affects the inference behavior of expert networks by visualizing the expert load within a specific layer from two perspectives: the number of tokens initially assigned to each expert (“Assigned Tokens”) and the number of tokens that pass the router and are actually executed (“Executed Tokens”).
Router-tuning prompts the experts to skip less important tokens and significantly lower the load of overloaded experts, which alleviates the imbalanced distribution of token assignments. Consequently, this approach fosters a more balanced utilization across the entire set of experts \cite{fedus2022switchtransformersscalingtrillion, he2025capacityawareinferencemitigatingstraggler} and thus enhances the efficiency.

\paragraph{Integration with LoRA} 
Router-Tuning enables dynamic depth to improve computational efficiency, whereas parameter-efficient fine-tuning (PEFT) methods aim to update a small subset of parameters to enhance downstream task performance. To examine whether Router-Tuning is complementary to PEFT, we propose jointly conducting Router-Tuning with LoRA fine-tuning~\cite{hu2021loralowrankadaptationlarge}, targeting improvements in both efficiency and task performance. As shown in Table~\ref{tab:mod_lora}, this joint training strategy preserves the efficiency benefits of Router-Tuning while maintaining the performance gains achieved by LoRA. Together, the integration of Router-Tuning and LoRA offers a more advanced fine-tuning paradigm that further enhances overall model capability.

%% file: sections/Limitations.tex
\section*{Limitations}
\label{:sec:Limitations}
Despite the progress achieved in this work, several limitations remain. First, while we have advanced MoD through Router-Tuning, other, potentially more sophisticated training strategies may further improve performance and merit future investigation. Second, due to computational resource constraints, our experiments were limited to a small set of models and tasks. Extending this approach to a broader range of architectures and applications would provide deeper insight into its generalizability and full potential.